\documentclass[11pt, twocolumn]{article}
\usepackage[margin=1in]{geometry}
\usepackage{graphicx} 
\usepackage{amsmath} 
\usepackage{hyperref} 
\usepackage{float} 
\usepackage{subcaption} 
\usepackage{authblk} 
\usepackage{listings} 
\usepackage{color} 
\usepackage{tikz} 
\usepackage{multirow} 
\usepackage{tabularx} 
\usepackage{todonotes}
\usepackage{authblk}
\usepackage{hyperref}  
\usepackage{cuted}     

\title{Automated and Holistic Co-design of Neural Networks and ASICs for Enabling In-Pixel Intelligence}

\author[1]{Shubha R. Kharel\thanks{\href{mailto:skharel@bnl.gov}{skharel@bnl.gov}}}
\author[2]{Prashansa Mukim}
\author[2]{Piotr Maj}
\author[2]{Grzegorz W. Deptuch}
\author[1]{Shinjae Yoo}
\author[1]{Yihui Ren}
\author[2]{Soumyajit Mandal}

\affil[1]{Computational Science Initiative, Brookhaven National Laboratory, Upton, NY 11973}
\affil[2]{Instrumentation Department, Brookhaven National Laboratory, Upton, NY 11973}

\date{}

\begin{document}

\maketitle

\begin{abstract}
    Extreme edge-AI systems, such as those in readout ASICs for radiation
    detection, must operate under stringent hardware constraints such as
    micron-level dimensions, sub-milliwatt power, and nanosecond-scale speed
    while providing clear accuracy advantages over traditional architectures.
    Finding ideal solutions means identifying optimal AI and ASIC design
    choices from a design space that has explosively expanded during the merger
    of these domains, creating non-trivial couplings which together act upon a
    small set of solutions as constraints tighten. It is impractical, if not
    impossible, to manually determine ideal choices among possibilities that
    easily exceed billions even in small-size problems. Existing methods to
    bridge this gap have leveraged theoretical understanding of hardware to
    create proxies for key metrics such as ASIC area and power and used them in
    neural architecture search. However, the assumptions made in computing such
    theoretical metrics are too idealized to provide sufficient guidance during
    the difficult search for a practical implementation. Meanwhile, theoretical
    estimates for many other crucial metrics (like delay) do not even exist and
    are similarly variable, dependent on parameters of the process design kit
    (PDK). To address these challenges, we present a study that employs
    intelligent search using multi-objective Bayesian optimization, integrating
    both neural network search and ASIC synthesis in the loop. This approach
    provides reliable feedback on the collective impact of all cross-domain
    design choices. We showcase the effectiveness of our approach by finding
    several Pareto-optimal design choices for effective and efficient neural
    networks that perform real-time feature extraction from input pulses within
    the individual pixels of a readout ASIC.
\end{abstract}

\section{Introduction and Motivation}

Real-time edge-AI systems are designed for inference at the edge of computing
hardware, where data is generated, and are crucial for applications requiring
swift detection, inference, and
decision-making~\cite{li2019edge,merenda2020edge,wang2020edge}. These systems
must be adequately reliable while operating within stringent resource
constraints, including limited area, power, and latency. Their design
complexity is amplified by the numerous design choices available in both the AI
and hardware spaces, the complex interactions between them, and their combined
effect on the deployment feasibility in real-world
applications~\cite{mazumder2021survey}.

Tackling the stringent requirements for extreme edge-AI systems has spurred
innovations in AI models, hardware, and their co-design. In AI design, research
has concentrated on model compression techniques such as quantization and
pruning~\cite{liang2021pruning,hawks2021ps} that enhance computational
efficiency without detrimentally compromising accuracy. Similarly, innovations
have emerged in AI accelerators customized for these AI models and their
implementation on field-programmable gate arrays
(FPGAs)~\cite{gao2020edgedrnn,sipola2022artificial} and application-specific
integrated circuits (ASICs)~\cite{houshmand2022diana,prabhu2022chimera}. The
co-design approach has taken this process further by simultaneously considering
design choices from both AI and hardware perspectives to find the best solution
for both domains~\cite{bringmann2021automated,zhang2022algorithm}.

This paper focuses on the co-design approach to addressing the challenges of
extreme edge-AI systems. A significant challenge in co-design is the
exponential growth in design choices and the substantial compute costs required
to quantify its impact on design objectives. Automated searches with advanced
algorithms like Bayesian optimization~\cite{reagen2017case,shi2020using} and
reinforcement learning~\cite{abdelfattah2020best,jiang2020hardware} have shown
promise in navigating this complex co-optimization space but remain too
expensive to be practical for most applications, especially when quantifying
the costly hardware objectives. Some progress has been made using theory-guided
approximations to model hardware
objectives~\cite{coelho2021automatic,campos2023end}, but actual implementations
often deviate significantly from these estimates, leading to sub-optimal or
infeasible designs. While improving the accuracy of these estimates is crucial,
it is often insufficient due to the unique and intricate nature of more
practical hardware objectives for which theoretical estimates are not
available.

Many scientific applications would benefit from fast, low-latency machine
learning, including event reconstruction and triggering in particle
accelerators, denoising of gravitational wave detectors, and real-time
monitoring and control of plasma
dynamics~\cite{govorkova2022autoencoders,deiana2022applications}. ASICs
represent a natural progression from the success of FPGA-based AI accelerators
in such scientific applications. Unlike other specialized hardware, ASICs offer
a unique combination of efficiency, area, and power optimization, enabling them
to operate under the extreme conditions of edge computing and even advancing AI
to pixel-level processing in devices such as radiation
detectors~\cite{miryala2022design,miryala2022peak,miryala2022waveform,mandal2023real}.
Despite these advantages, the field has remained relatively unexplored due to
the high development costs, the need for close interdisciplinary collaboration
with highly specialized hardware design domains - often lacking in academic and
scientific settings - and the paucity of consumer-available options. However,
the growing availability of open-source tools, increased support from
government initiatives, and heightened scientific interest in advancing
efficient AI has created a more feasible ecosystem for AI-ASIC co-design.

The AI-ASIC co-design environment presents more constraints than those
typically encountered with AI accelerators and FPGAs. While this adds
complexity, it also narrows the design space, enabling a more feasible approach
for holistic integration, which is often challenging in broader settings. In
this study, we embraced this unique context by holistically integrating design
choices across the AI and ASIC domains. Instead of relying on theoretical
estimates, our approach utilizes direct metrics obtained from hardware
synthesis, incorporating these insights into a fully automated pipeline using
multi-objective Bayesian optimization. This method allows us to consider more
practical and necessary hardware objectives. We demonstrate the effectiveness
of our approach on a scientific application, namely the design of on-chip
neural networks for feature extraction from amplified charge
pulses~\cite{miryala2022waveform}. Such pulses, which are produced by readout
ASICs designed for radiation sensing, encode information on the energy,
direction of arrival, and identity of particles observed in high-energy physics
experiments~\cite{radeka2011signal}.

The main contributions of this work are as follows:

\begin{itemize}
    \item \textbf{Realistic Estimates and Practicality:} We incorporate more accurate estimates of design objectives in our optimization process to increase the likelihood of achieving designs that are both optimal and feasible.

    \item \textbf{Holistic Co-design integration:} We expand on the usual co-design choices, such as the neural network architectures and quantization strategies, to also simultaneously consider ASIC synthesis strategies, thus enabling a more comprehensive search for practical designs.

    \item \textbf{Diversity of Co-design Objectives:}  We also expand our design objectives to include practically useful metrics such as the time delay of the circuit implementation, in addition to traditional optimization goals like accuracy, area, and power.

    \item \textbf{Open-source automation tool} We provide the scientific community with an automated pipeline that utilizes fully open-source tools from both AI and ASIC domains. We also make data from thousands of ASIC synthesis runs available for further research.
\end{itemize}

Existing applications of AI algorithms in radiation detection are focused on offline data analysis, such as for event clustering and track reconstruction~\cite{MLinPP1,MLinPP2}. These algorithms are generally implemented on graphics processing units (GPUs), which offer reconfigurability at the cost of non-deterministic latency and low energy efficiency. Thus, they are unsuitable for emerging applications, such as real-time denoising, that require online operation of the AI algorithms. The hardware platform of choice for such real-time applications has traditionally been FPGAs, which combine bit-level reconfigurability with deterministic latency. However, the energy efficiency of FPGAs, as measured using a suitable metric such as energy-delay product (EDP), is generally higher than that of GPUs but suffers from the significant overhead required to support the programming fabric. Thus, recent work has proposed AI accelerators implemented on the ``edge'' close to the data source, i.e., within the readout ASICs used for amplification and conditioning of charge waveforms~\cite{di2021reconfigurable,miryala2022waveform}. Such ASIC implementations offer significantly higher energy efficiency than FPGAs while also allowing the design to be optimized for the extreme environmental conditions (e.g., cryogenic or high-radiation) often encountered in near-detector environments. However, these advantages come at the cost of limited post-fabrication reconfigurability. Given the limited power, area, and computational resources of the readout ASICs, pre-fabrication optimization of the AI model hyperparameters thus becomes critical for successful implementation of the readout system.

Typical readout ASICs are arranged in either 1-D arrays (known as strip detectors) or 2-D arrays (known as pixelated detectors). Quantitative limits on these metrics depend on the characteristics of the detector and experiment. For example, the ALICE ITS3 large-area silicon vertex tracker (SVT) uses small pixels ($15$~$\mu$m $\times 15$~$\mu$m in size) with binary readout~\cite{rinella2023digital,groettvik2024alice}, which results in strict power constraints ($<100$~nW/pixel). On the other hand, full-field X-ray fluorescence detectors use larger pixels (typically $100$~$\mu$m $\times 100$~$\mu$m in size) with high-resolution amplitude readout~\cite{Gorni_2024}, which results in somewhat relaxed power constraints ($<50$~$\mu$W/pixel). In general, the area and power requirements for strip detectors are further relaxed compared to those for pixels. Thus, they were the first targets for implementing on-chip edge AI, as in our previous work~\cite{miryala2022design,miryala2022peak,miryala2022waveform}. Latency requirements also vary significantly depending on the average particle arrival rate, but typically range from 0.1-100~$\mu$s.

\section{Related Work}

A common metric used to characterize the performance of digital systems the power-delay product (PDP), often expressed in inverse form as performance-per-watt (MIPS/W or FLOPS/W) with higher values being better~\cite{brooks2000power}. Another common metric is the EDP, which is defined as the product of delay/operation and energy/operation~\cite{horowitz1994low} with lower values being better. The two metrics are related by the fact that energy is proportional to PDP, which in turn implies that EDP is proportional to (performance)$^2$-per-watt, i.e., emphasizes performance to a greater degree than PDP.

Earlier work on automated hardware-aware design space exploration for AI utilized analytical models to approximate the relationship between network hyperparameters (such as the number of layers, neurons per layer, and weight precision per layer) and performance metrics such as PDP. For example, QKeras~\cite{coelho2021automatic} utilizes an analytical energy model of the form
\begin{equation}
    E_{layer} = E_{input} + E_{param} + E_{MAC} + E_{output},
    \label{eq:qkeras_energy}
\end{equation}
where $E_{layer}$, $E_{input}$, $E_{param}$, $E_{MAC}$, and $E_{output}$ denote the energy consumed by a given layer, to read the outputs and parameters, to perform multiply-accumulate (MAC) operations, and store the outputs, respectively. Models for each term are derived from empirical data in a 45~nm CMOS process listed in~\cite{horowitz20141}. While computationally efficient, the accuracy of this analytical approach is limited since it does not take the actual circuit structure into account. Thus, the analytical models do not account for design optimizations performed during physical implementation, including the processes of logic synthesis, timing analysis, and placement \& routing. In this paper, we obtain more accurate performance estimates by including logic synthesis within our Bayesian optimization framework.

\section{Methodology}

\begin{figure}[t]
    \includegraphics[width=\linewidth]{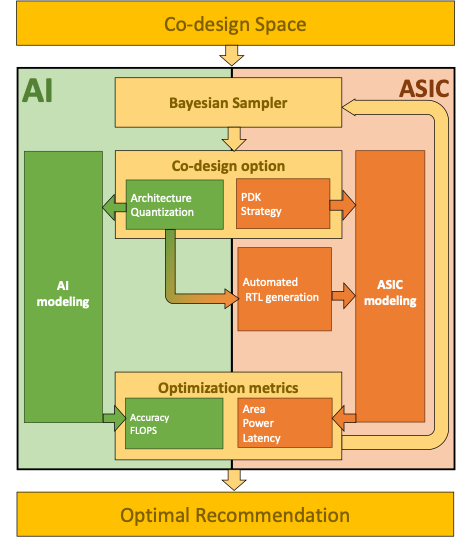}
    \caption{Overview of methodology.}
    \label{fig:overview}
\end{figure}

In our co-design method, we collate the Neural Network and ASIC design spaces
and simultaneously optimize multiple objectives, using realistic metrics
affected by a given set of choices. The Neural network design space consists of
MLP architecture and heterogeneous quantization configuration. ASIC design space
consists of the strategy used during ASIC synthesis. ASIC synthesis and
Quantization Aware training within it gives a more accurate estimate of
co-design objectives. Finally, an automated pipeline integrates all components
to do multi-objective Bayesian optimization. This method keeps things simple,
suitable, and sufficient for demonstrating our co-design approach.

\subsection{Neural Network Architecture}

We focus on Multi-Layer Perceptron (MLP) models consisting of fully connected
layers. Despite their simplicity, they have been proven to be universal
approximators, allowing them to approximate the most relevant functions with
sufficient depth and width, both of which are included in the co-design space.
While these parameters determine the model's performance, they also determine
the computational cost which scales as \(O_{\text{MLP}} = O\left(\sum_{i=1}^{d}
n_i \cdot n_{i+1}\right)\), where \(d\) is the depth and \(n_i\) is the width
of each layer. Computational complexity affects relevant co-design objectives
by influencing the area needed for memory and computation units, the power
consumed for data storage and processing, and the latency required to compute
each layer's output.

\subsection{Quantization Configuration}

Quantization reduces the precision of the numerical representation of weights
and activation functions from larger floating-point numbers to smaller fixed-point
numbers, integers, or even binary numbers. Quantization can significantly
decrease resource consumption for an MLP with a given width and depth without
substantially affecting its performance. Heterogeneous quantization takes this
idea further by aggressively quantizing less important layers to lower
precision based on the importance of the layer to the overall performance in
the network. So, we extend the co-design space of the neural network to include
the layer-specific quantization configuration for  MLP with given depth and
widths.

\subsection{ASIC Design}

Recent years have witnessed a rapid growth in the availability of open-source process design kits (PDKs) and electronic design automation (EDA) tools for ASIC design. The first open-source PDK, for the SkyWater 130~nm CMOS process, was launched in 2020 in collaboration with Google. It has since been widely used for a variety of ASIC designs~\cite{edwards2021real}, including processors~\cite{zhu2022greenrio}, coarse-grained reconfigurable arrays (CGRAs)~\cite{chen2023open} and AI accelerators~\cite{modaresi2023openspike,parmar2023open}.

The standard open-source digital ASIC design flow for the SkyWater 130~nm process is OpenLANE, which is based on the OpenROAD project~\cite{ajayi2019toward} and was first introduced by Shalan and Edwards in 2020~\cite{shalan2020building}. OpenLANE has been benchmarked against commercial EDA tools~\cite{zhu2022greenrio} and found to be comparable on key metrics such as synthesis run time, gate count, placement \& routing time, die area, leakage power, placement density, and maximum clock frequency. However, its performance is typically $1.4\times$ to $1.6\times$ worse for most of these metrics. On the plus side, OpenLANE is highly automated, thus making it suitable for incorporation within Bayesian design space exploration.

\subsection{Multi-objective Bayesian Optimization}

The most limiting aspects of co-designing AI and ASIC are the high
computational cost, exponential growth in design choices, and coupled competing
objectives. The computing cost is even exacerbated when we include ASIC
synthesis to get more accurate estimates of the ASIC objective metrics. Naively
exploring the exponential and costly design space, like grid search, is
inefficient, if not infeasible. To handle these challenges, we use Bayesian
optimization, designed to minimize the number of expensive trials and proven to
work effectively in domains including machine learning hyper-parameter tuning,
algorithm selection, and control systems optimization. Being a black-box
optimization, it also makes it easy to integrate and extend for further
research. Additionally, it can be extended to globally optimize for multiple
conflicting objectives, like accuracy and efficiency, in our co-design problem.
The result of this optimization process gives us Pareto-optimal solutions that
balance accuracy/efficiency trade-offs, enabling us to choose from a set of
optimal designs of AI and ASIC.

\subsection{Integration and Automation}

While plenty of tools are available for NN modeling and some for ASIC
synthesis, there is a high barrier between tools from both co-design domains.
Our work bridges this much-needed gap through the seamless integration of these
tools. Additionally, automation for going back and forth between NN training
and ASIC synthesis is also lacking despite being crucial for co-design. Our
work contributes to this by fully automating co-design optimization using the
integrated tools. To make our tools accessible for future research, we
exclusively use open-source tools.

\section{Implementation}

Our goal is to provide the research community with automated, open-source,
extensible, and scalable tool integration that is familiar to researchers in
both AI and ASIC domains. With this in mind, our automated pipeline uses Optuna
for Bayesian optimization, QKeras for quantization-aware training, OpenLANE for
ASIC synthesis, Hydra for easy configuration management, and Docker for
containerization.

Optuna is a popular hyperparameter optimization library that offers features
like parallelization and multiple optimization algorithms and is already
popular among ML researchers. QKeras provides quantization-aware training for
heterogeneously quantized models and is among the first few open-source
libraries to do so. OpenLANE is an open-source ASIC synthesis tool that
provides a complete RTL-to-GDSII flow. Hydra is implemented to allow easy
switching between different Bayesian optimization algorithms, specifying
co-design space and optimization objectives. Dockerization of both QKeras and
OpenLANE enables easy scalability through the distributed parallelization that
is well-integrated into Optuna.

The process proceeds as follows: Optuna suggests the MLP architecture,
quantization configuration, and ASIC synthesis strategy for the subsequent
trials based on the results of all previous choices and their effects on all
optimization objectives. Each suggested trial is then handled by an
'orchestrator' module that runs two containerized 'explorer modules': (a) the
NN Training module, which performs quantization-aware training using QKeras,
following the suggested MLP architecture and quantization configuration and
returns the validation loss of training; and (b) the ASIC Synthesis module,
which automatically generates the RTL code for the suggested MLP architecture
and quantization configuration using an easy-to-extend abstract syntax tree
abstraction, and then runs the ASIC synthesis using OpenLANE to report on ASIC objectives like area, power, and delay by parsing the synthesis reports.

\section{Experiment}

As an example of practical importance, here we consider the problem of on-chip waveform processing for radiation detectors. The basic goal of waveform processing is real-time feature extraction from pulse shapes. The latter are generated by analog front-end (AFE) circuits that amplify and filter the charge packets generated by energetic particles~\cite{radeka2011signal}. A generic pulse can be written in the form
\begin{equation}
    s(t) = Ap(t-t_0) + n(t)
\end{equation}
where $p(t)$ is a reference pulse shape, $A$ is a scaling factor known as the pulse amplitude, $t_0$ is the time of arrival (TOA), and $n(t)$ is additive noise. Note that $A$ is typically proportional to particle energy, so quantifying its value is a key goal of many measurements (which are known as amplitude spectroscopy). In other words, $A$ is the most important feature to be extracted from the waveform shape, $s(t)$. The TOA is also important in many cases since it encodes information on time of flight or arrival angle. Similarly, the reference shape $p(t)$ can provide additional information, such as particle identification (PID) or the angle of arrival.

For simplicity, we assume that the feature extraction is performed digitally. In this case, the AFE is followed by an analog-to-digital converter (ADC) that converts $s(t)$ to the discrete-time sequence
\begin{equation}
    s[k] = Ap[k-k_0] + n[k]
\end{equation}
where $k$ denotes the sample index. The digitized reference shape, $p[k]$, depends on the particle type, detector material, and AFE filtering parameters. A commonly used model for $p[k]$ is the so-called CR-(RC)$^N$ pulse shape, which is given by
\begin{equation}
    p[k] = \frac{1}{{N!}}{\left( {\frac{kT}{\tau }} \right)^N}\exp \left( { - kT/\tau } \right)
    \label{eq:cr_rc}
\end{equation}
where $T$ is the sampling period, $N=1,2,\ldots$ is known as the order of the shaping function, and $\tau$ is the decay time constant.

\begin{figure}[t]
    \centering
    \includegraphics[width=0.80\columnwidth]{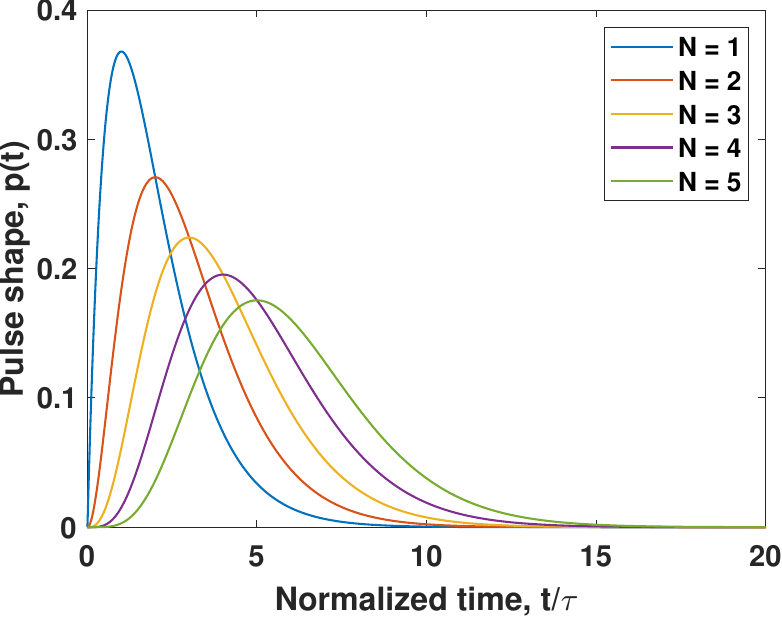}
    \caption{Reference shape, $p(t)$, of CR-(RC)$^N$ pulses for different values of $N$.}
    \label{fig:cr_rc_shaper_impulse_response}
\end{figure}

Fig.~\ref{fig:cr_rc_shaper_impulse_response} plots $p(t)$ in continuous time for various values of $N$. Note that the waveform becomes smoother and more symmetric as $N$ increases due to the increased amount of low-pass filtering. The peak value of $p(t)$, which occurs at $t_{pk}=N\tau$, is given by
\begin{equation}
    A_{pk}(N) = \frac{(N)^{N}}{N!}e^{-N}.
    \label{eq:apk_shaper}
\end{equation}
The value of $A_{pk}(N)$ decreases monotonically with $N$: from $\approx 0.37$ for $N=1$ to $\approx 0.18$ for $N=5$. Also, the time at which the peak response occurs, $t_{pk}$ (also known as the peaking time), is proportional to $N$.

The quality of the received signal is quantified by the peak signal-to-noise ratio (PSNR), which is defined as
\begin{equation}
    \mathrm{PSNR} = \frac{A_{pk}}{\sigma_n},
\end{equation}
where $\sigma_n$ is the standard deviation of the additive noise.

\subsection{Dataset and Models}

\begin{figure}[h]
    \includegraphics[width=\linewidth]{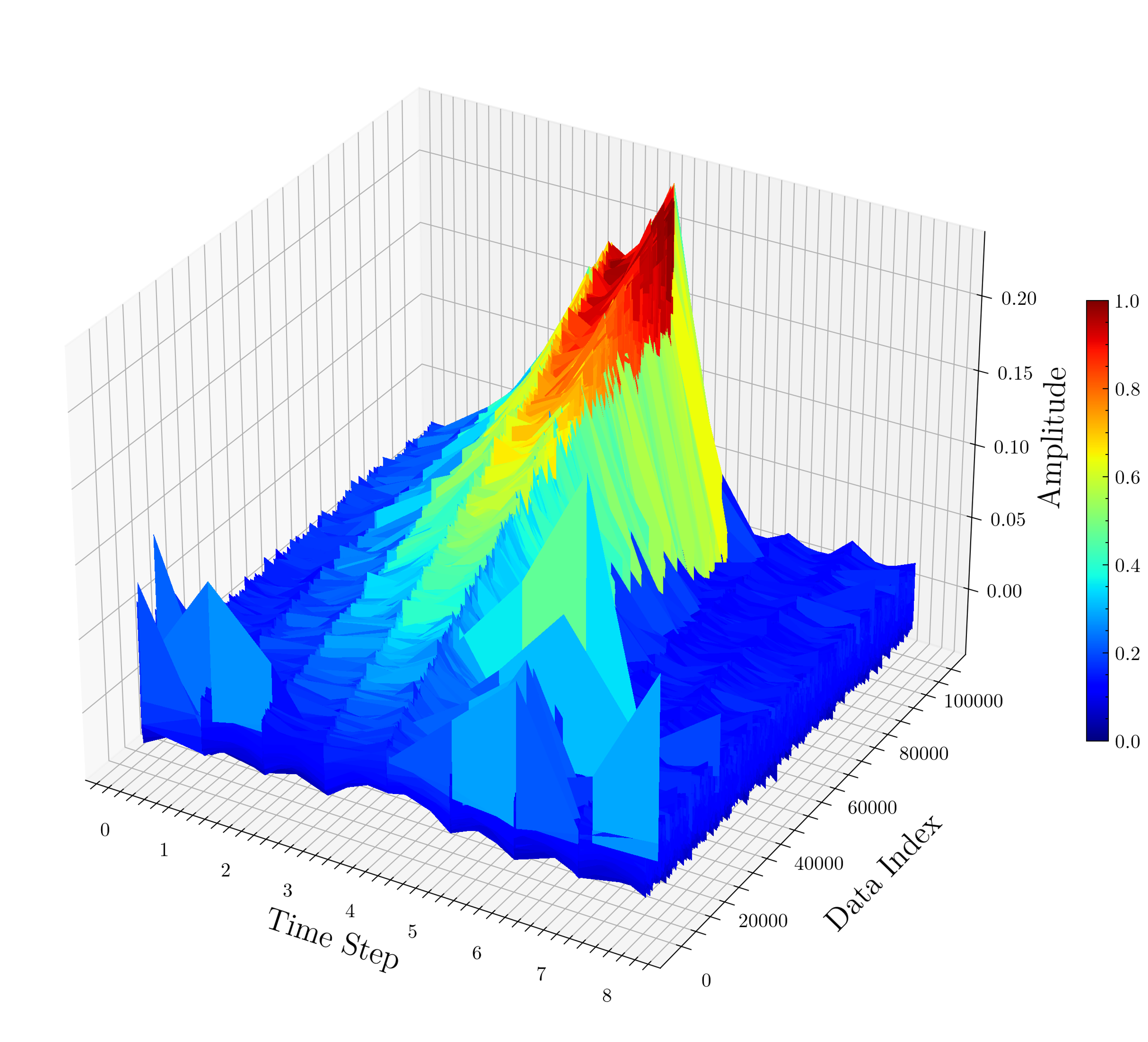}
    \caption{View of the 98,583 waveforms used in the experiment. Surface plot is made after
        sorting the waveform by amplitude along along the data index axis.}
    \label{fig:data}
\end{figure}

A set of noisy simulated pulses with the CR-(RC)$^n$ shape shown in
Eqn.~(\ref{eq:cr_rc}) was used for training. The noise component was assumed to
be additive, white, and Gaussian (AWG). The peak amplitude varied randomly
between pulses with a uniform distribution, resulting in a realistic PSNR range
of 1 to 20 (0~dB to 26~dB). Pulse arrival times were randomly distributed
following a Poisson distribution, as commonly encountered in experiments.
Datasets with different values of $N$ (varying between 1-5) were used to study
the impact of pulse shape changes.

Initial experiments were focused on datasets where both the pulse shape,
$p[k]$, and the time constant, $\tau$, were kept fixed. The feature extraction
process then reduces to the single-variable problem of estimating the pulse
amplitude, $A$. An optimal solution to this problem exists when the noise is
AWG in nature; it takes the form of a classical matched filter followed by peak
picking~\cite{mandal2023real}. However, note that this solution does not
account for the inevitable overlap between nearby pulses, which is known as
pileup.

Later simulations also allowed the time constant, $\tau$, to vary randomly over
the range $[\tau_0,2\tau_0]$, where $\tau_0=T$ is a constant. In this case, the
feature extraction algorithm is asked to estimate both $A$ and $\tau$, i.e.,
has to solve a two-variable problem.

The sampling rate for the simulations was initially set to $T=\tau$. All processing was performed on data windows with $N_W=32$ data points. Variations in pulse arrival times within these windows, which are inevitable due to the asynchronous pulse arrivals, was minimized by a real-time waveform alignment algorithm. This algorithm operates as follows:
\begin{itemize}
    \item Buffer incoming data into windows of length $2N_W$.
    \item Find the location within the buffer, $k_{max}$, with the peak absolute value, $|s[k]|$.
    \item Output the sub-window of length $(N_W+1)$ centered around the peak position, $k_{max}$.
    \item Buffer the next set of points and repeat.
\end{itemize}
Note that this algorithm makes no attempt to mitigate the effects of noise and pileup on $|s[k]|$. Thus, it only performs coarse time alignment of the received pulse waveforms.

Finally, the size of the neural network used for feature extraction was
minimized by decimating the time-aligned waveforms, i.e., by increasing the
sampling period from $T$ to $MT$ where $M>1$ is an integer. This process
reduces the length of each data window from $N_W+1$ to $N_W/M+1$.

Our experiments used a default decimation factor of $M=4$ during the
pre-processing stage, which results in windows containing 9 time points. Note
that decimation also makes the estimation problem more challenging. The dataset
was then split into training, validation, and test sets with a distribution of
70\%, 20\%, and 10\%, respectively. This resulted in 69,008 samples for
training, 19,716 for validation, and 9,859 for testing, ensuring comprehensive
coverage across different phases of model evaluation

\subsection{Co-design Space}

The co-design choices for the experiment includes: a) the number of hidden
layers, b) the number of perceptrons in each hidden layer, c) quantization
levels of weights/biases between layers, d) input/output quantization levels
for the MLP, and e) ASIC synthesis strategies.  This makes the dimension of the
co-design space to be:

\begin{equation}
    C = N_{Q_{io}} \times N_{A_s} \times \sum_{d=1}^{D} (N_W \times N_Q)^d
\end{equation}

where \(N_{Q_{io}}\), \(N_{A_s}\), \(N_W\), \(N_Q\), and \(D\) are the number
of choices available for input/output quantization, ASIC synthesis strategies,
perceptrons in a layer, quantization levels in a layer, and the number of
hidden layers, respectively. The range of values used in the experiment and
consequently the dimension of the co-design space is shown in Table
\ref{tab:co_design_space}.

\begin{table}[h]
    \centering
    \begin{tabular}{|l|c|}
        \hline
        \textbf{Co-design Choices}   & \textbf{Range}                     \\
        \hline
        Hidden Layers                & 1-3                                \\
        Perceptrons in a layer       & 2-18                               \\
        Weights/Biases Quantizations & 2-16                               \\
        Input/Output Quantizations   & 2-16                               \\
        ASIC Synthesis Strategies    & 9 choices                          \\
        \hline
        \multicolumn{2}{|c|}{Dimension of co-design space: 2,247,298,425} \\
        \hline
    \end{tabular}
    \caption{Range of co-design choices used in the experiment}
    \label{tab:co_design_space}
\end{table}

\section{Results}

In this experiment, we defined co-design objectives, including the validation
loss, area, power, and delay estimates reported from the ASIC synthesis. Some
objectives, such as loss, area, and delay, are competitive, whereas others,
like area and power, mostly overlap. There is no optimal solution in
multi-objective optimization scenarios like this. Instead, a set of solutions
exists where each is Pareto optimal—improvement in one objective necessitates a
trade-off in another.

\subsection{Pareto Fronts}

\begin{figure}[t]
    \includegraphics[width=\linewidth]{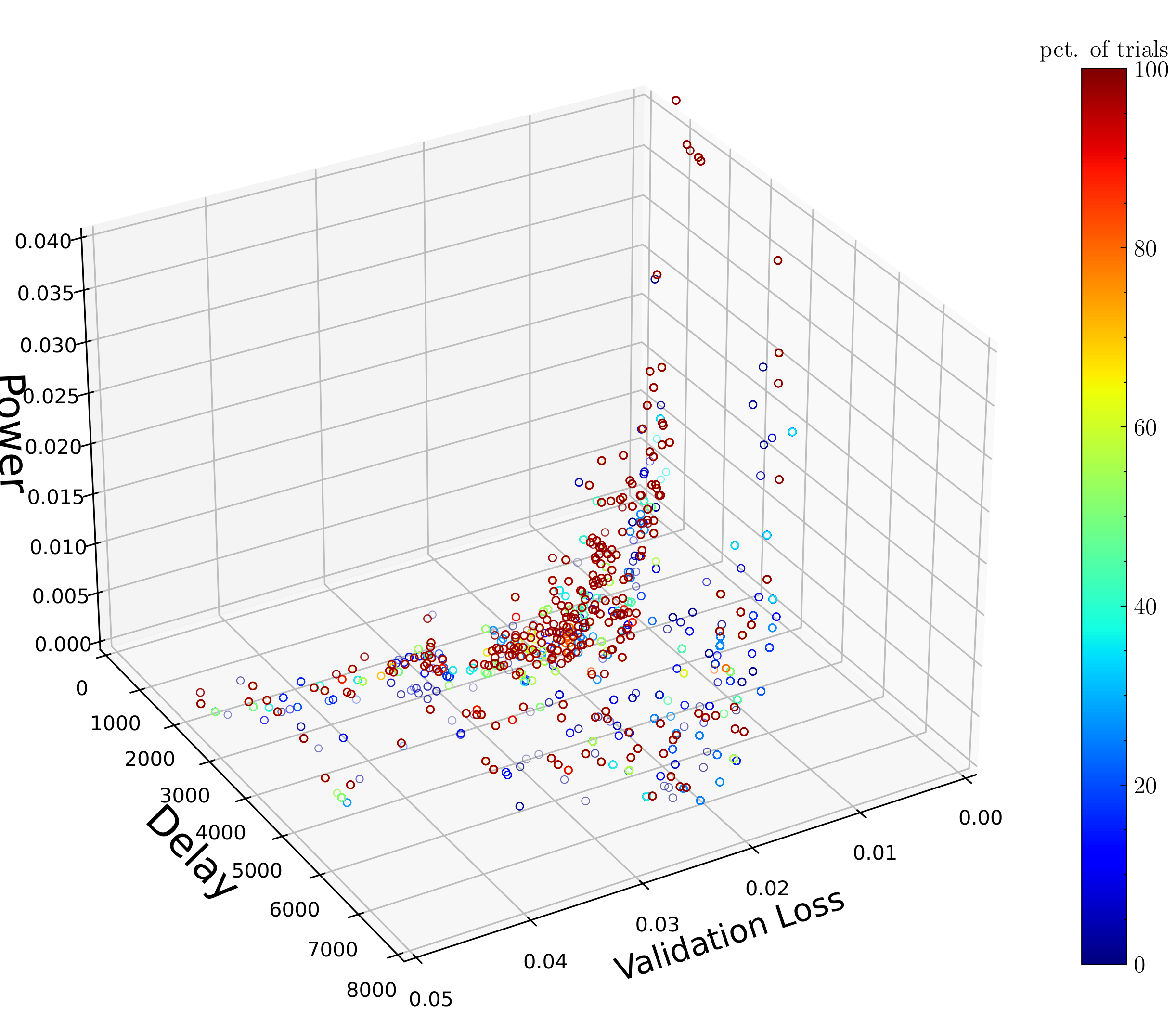}
    \caption{Evolution of Pareto Front during optimization.
        Color is based on iteration point where the Pareto belongs, in terms of percentage of total iterations. The size of the points is proportional to the area.
    }
    \label{fig:pareto_3d}
\end{figure}

\begin{figure}[t]
    \centering
    \includegraphics[width=0.9\linewidth]{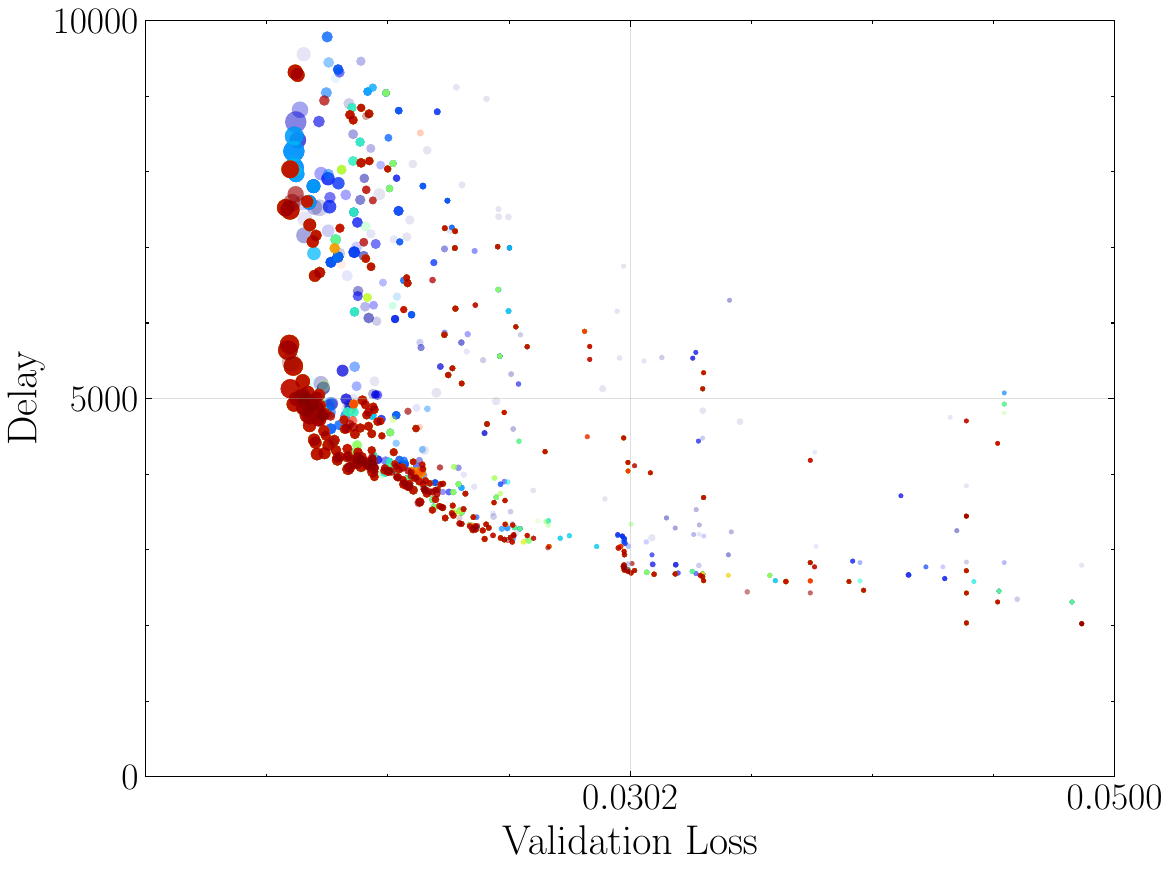}
    \caption{2D cross-sections of 3D Pareto front from Fig.~\ref{fig:pareto_3d}
        showing the relationship between validation loss and delay.}
    \label{fig:pareto2d_first}
\end{figure}

\begin{figure}[t]
    \centering
    \includegraphics[width=0.9\linewidth]{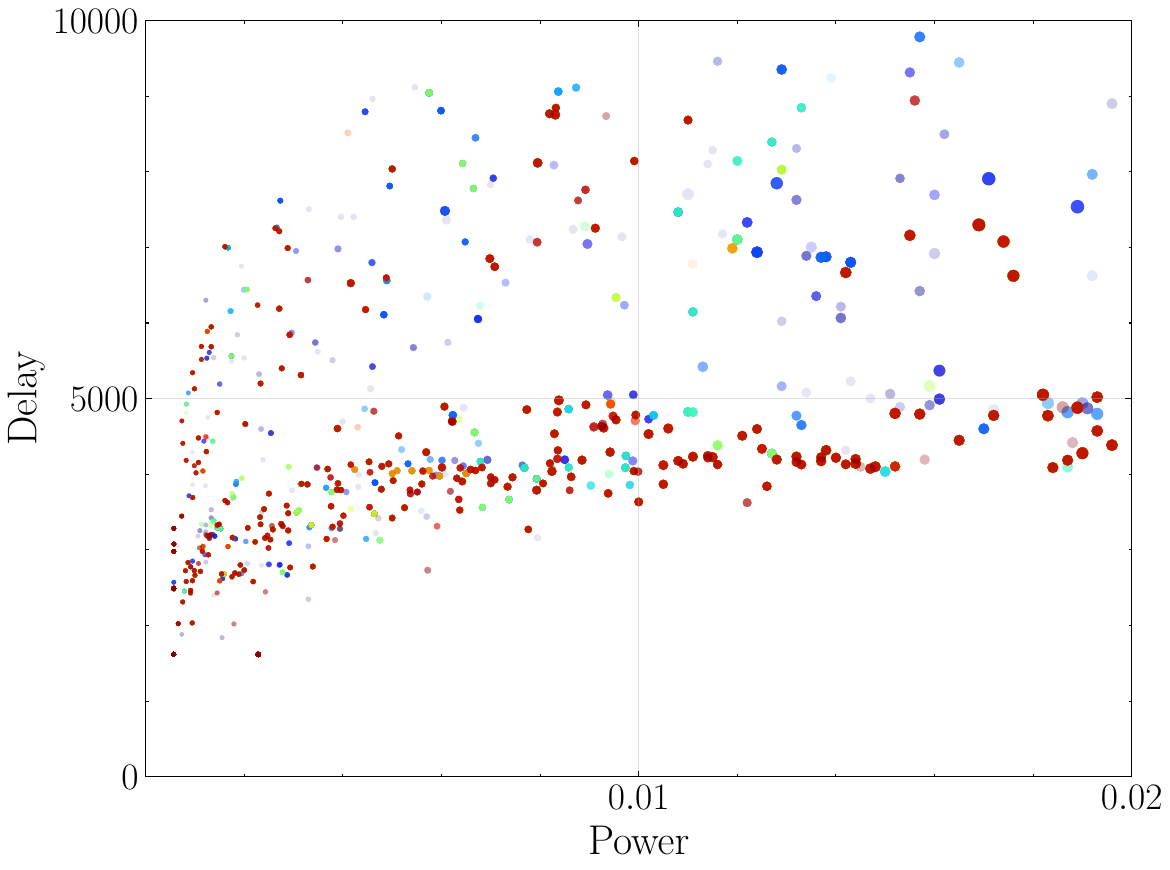}
    \caption{2D cross-sections of 3D Pareto front from Fig.~\ref{fig:pareto_3d}
        showing the relationship between power and delay.}
    \label{fig:pareto2d_second}
\end{figure}

\begin{figure}[t]
    \centering
    \includegraphics[width=0.9\linewidth]{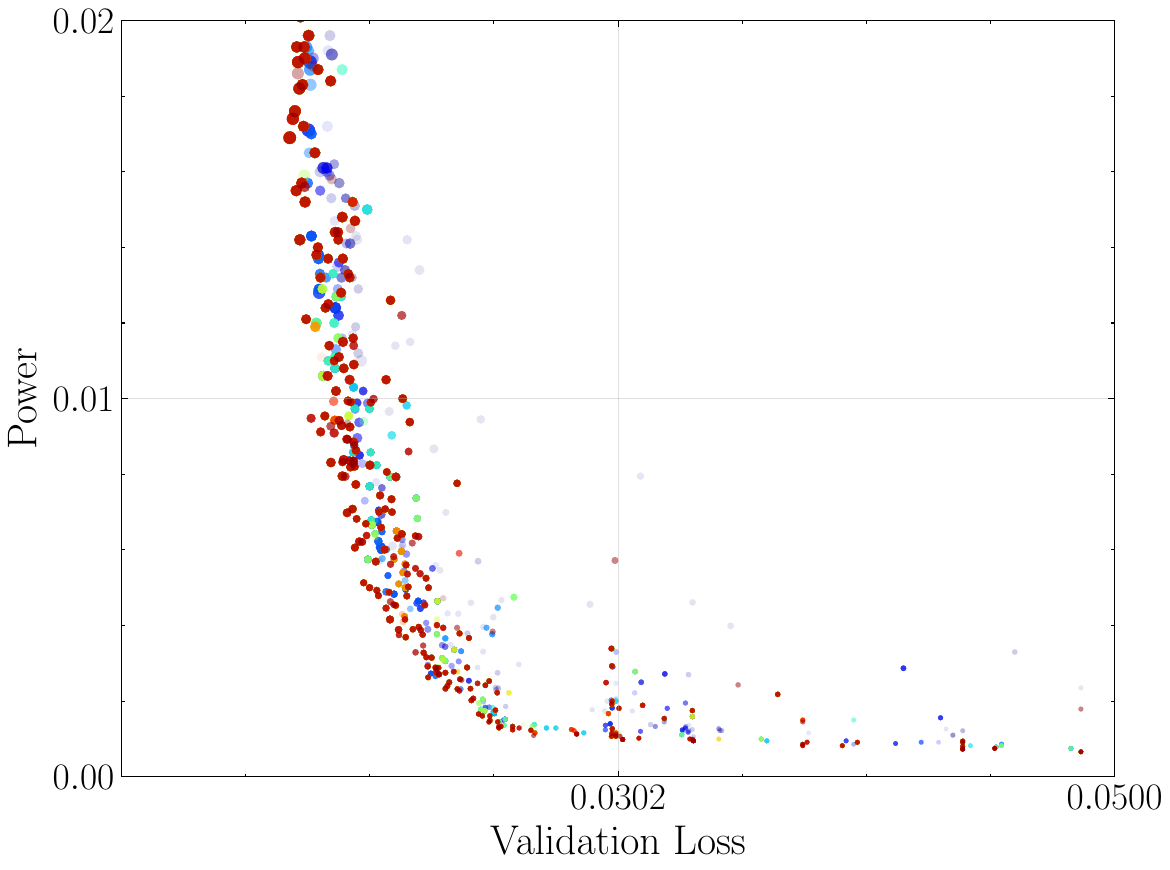}
    \caption{2D cross-sections of 3D Pareto front from Fig.~\ref{fig:pareto_3d}
        showing the relationship between validation loss and power.}
    \label{fig:pareto2d_third}
\end{figure}

The 3D cross-section inf Fig.~\ref{fig:pareto_3d} reveals some insight into the
search space explored during iterations. Early iterations, represented by blue
points, are dispersed widely across the graph, indicating an initial
exploration phase of the solution space. As the iterations advance, marked by a
shift toward red points, there is a noticeable clustering of data points,
suggesting that the optimization algorithm is converging toward a set of
optimal solutions and also reveals the topology of the front (we will continue
this discussion in a later section). Meanwhile, the topology of the
Pareto-front is also shown in those clusters, which brings us to the next set
of figures on 2D cross-sections (Figs.~\ref{fig:pareto2d_first} -
\ref{fig:pareto2d_third}), which gives more insight into the topology of the
front.

The 2D cross-section plots The most clear boundary appears when loss and power
are trying to be balanced. Fig.~\ref{fig:pareto2d_third} reveals a steep
relationship between validation loss and power, indicating significant
challenges in achieving lower validation losses. Points even from early
iterations densely cluster at lower validation loss levels, suggesting the
algorithm quickly reaches this boundary where further reductions in loss
require disproportionately higher power consumption. Moreover, the dense
clustering of data points along the curve suggests a robust and predictable
trade-off between these objectives. The topology of this Pareto front is
smoother and more continuous compared to the other cross-sections, indicating
that the optimization paths for reducing validation loss are closely tied to
increases in power usage. The narrow variance is likely the reason behind
previous success of loss-power optimization based on theoretical estimates
implemented in popular libraries seems to do well enough.

The delay-power boundary can be seen in Fig.~\ref{fig:pareto2d_second}, which
reveals a broad spread of data points, indicating a diverse range of outcomes.
Lower power can sometimes be achieved with high and low delays, further
complicating the trade-offs. This distribution suggests that finding
configurations that optimize one or both metrics might be possible without a
direct correlation between increased power and increased delay.

The Fig.~\ref{fig:pareto2d_third} also explores the relationship between delay
and validation loss. Like Fig.~\ref{fig:pareto2d_first}, there is a visible
concentration of points towards lower validation losses with varying degrees of
delay. Notably, the delay does not show a clear upward or downward trend with
decreasing validation loss, suggesting a more complex relationship where
specific configurations might optimize delay without significantly impacting
validation loss. Complex dynamics like this are one of the justifications of
our approach, in that some of the ASIC synthesis metrics are hard to estimate
using a theoretical approach.

\subsection{Optimization Dynamics}

\begin{figure}[t]
    \includegraphics[width=\linewidth]{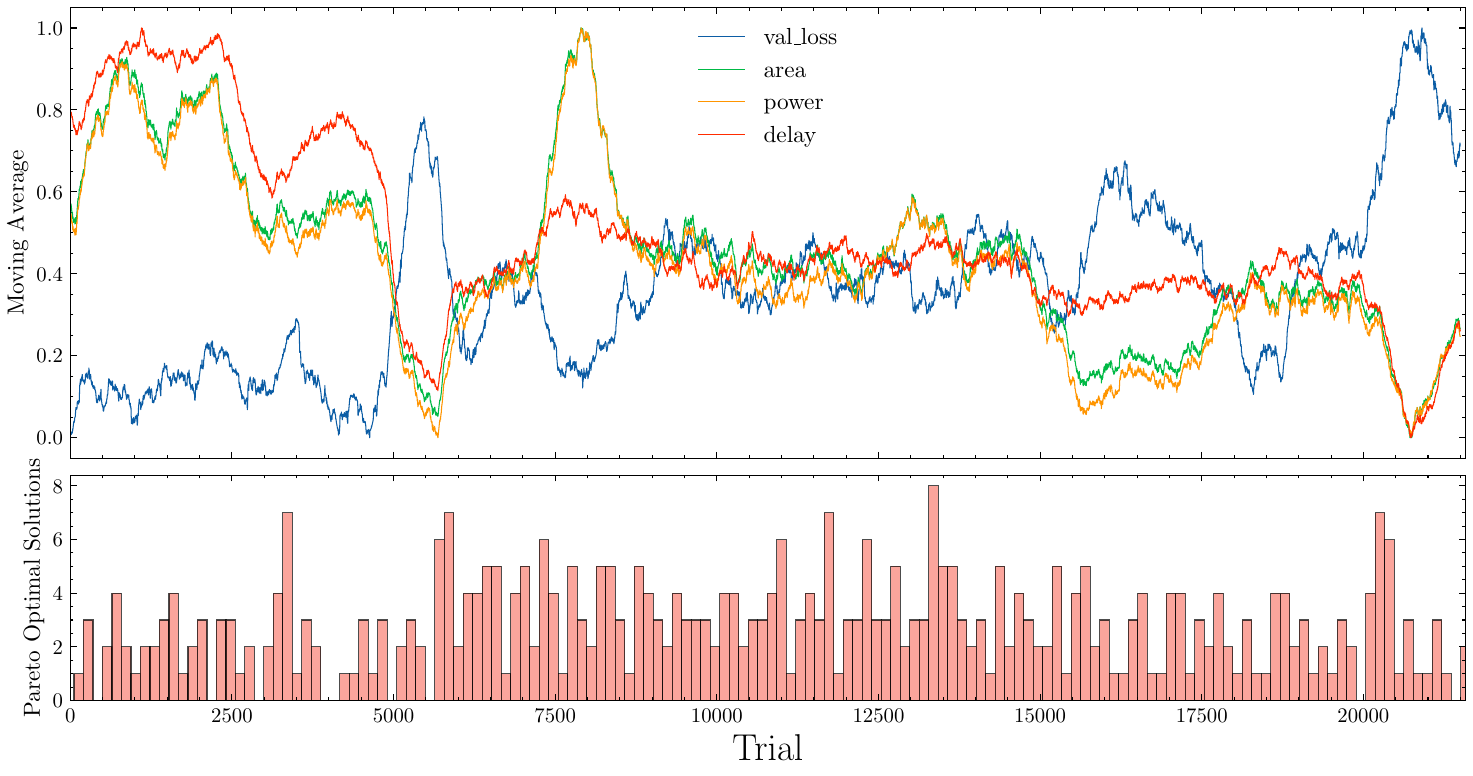}
    \caption{(top) Moving average of the objective values at each trail when
        optimization progresses (bottom) Number of optimal Final Pareto optimal
        points that came during iterations in that region.}
    \label{fig:obj_dynamics}
\end{figure}

\begin{figure}[t]
    \includegraphics[width=\linewidth]{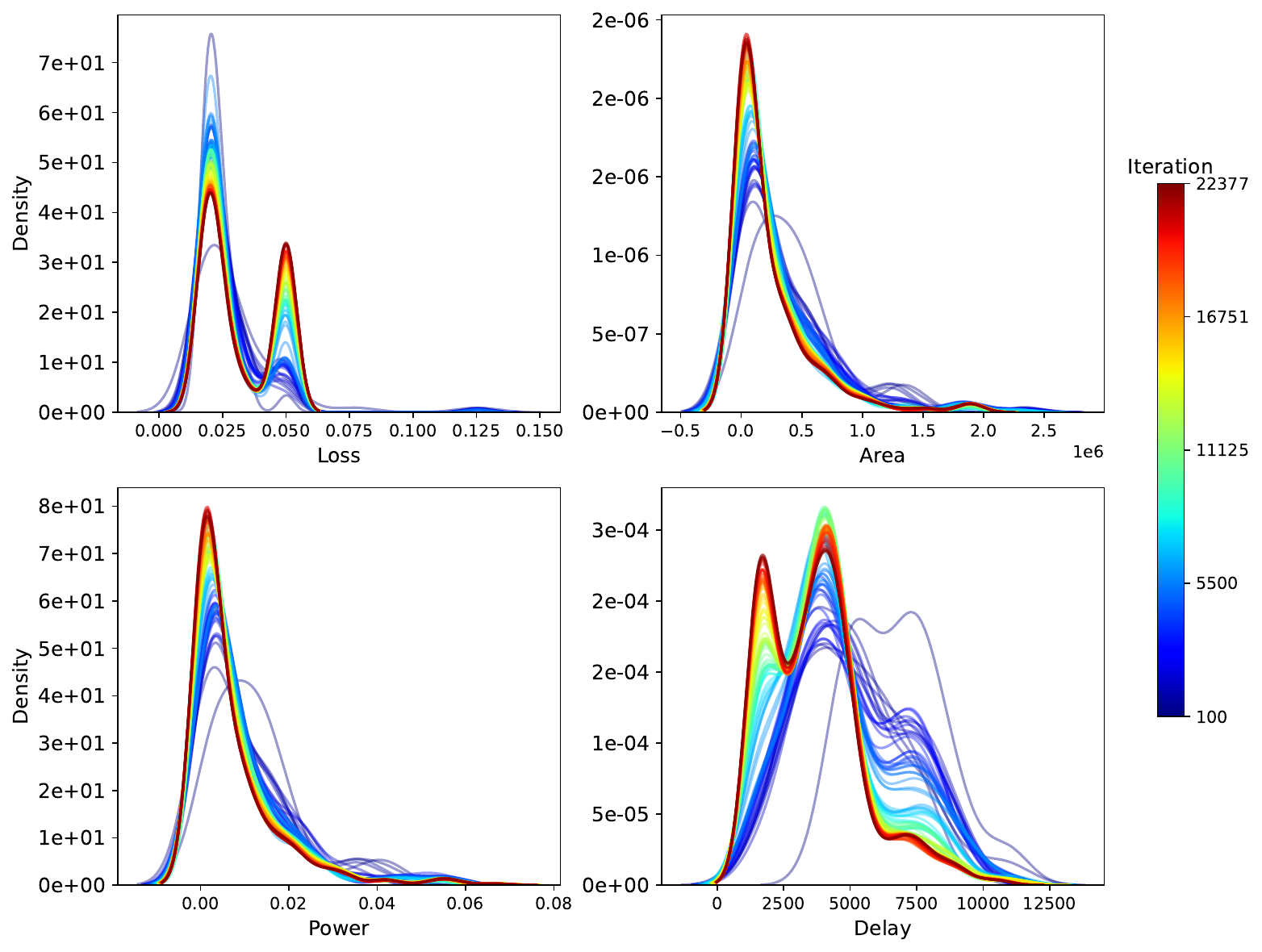}
    \caption{Kernel density estimate of objectives during optimization.}
    \label{fig:kde}
\end{figure}

Fig.~\ref{fig:obj_dynamics} shows the moving average of the optimization
objectives as the search progresses. The histogram shows the number of Pareto
optimal solutions during that evolution. The histogram shows that increasing
iterations consistently find Pareto optimal points throughout the optimization.
The Fig.~\ref{fig:kde} shows the kernel density estimate (KDE) of objectives in
Pareto fronts during different iterations.

There are several key observations from the optimization dynamics that can be
gained from Fig.~\ref{fig:obj_dynamics} and Fig.~\ref{fig:kde}. First, the
search increasingly goes from the region where loss is low to higher, apparent
in both Fig.~\ref{fig:obj_dynamics} and the top-left of Fig.~\ref{fig:kde}. The
second peak on the validation loss that is not being pushed further is also
close to the baseline validation from a model that always predicts the average
of the training set. This is happening because no constraint minimal
performance was set during this round of simulation. There is a high
sensitivity of area/power with the validation loss, whereas delay is less
susceptible to it, both of which were also hinted previously by
Figs.~\ref{fig:pareto2d_first}-\ref{fig:pareto2d_third}. Kernel density
estimates from Fig.~\ref{fig:kde} indicate a successful focusing of the search
towards promising regions, with area/power being most challenging for the
optimization to push lower. The dynamics of delay seen in both
Fig.~\ref{fig:obj_dynamics} and Fig \ref{fig:kde} clearly show that it
benefited the most from the higher iteration of optimization.

\subsection{Convergence}

\begin{figure}[t]
    \includegraphics[width=\linewidth]{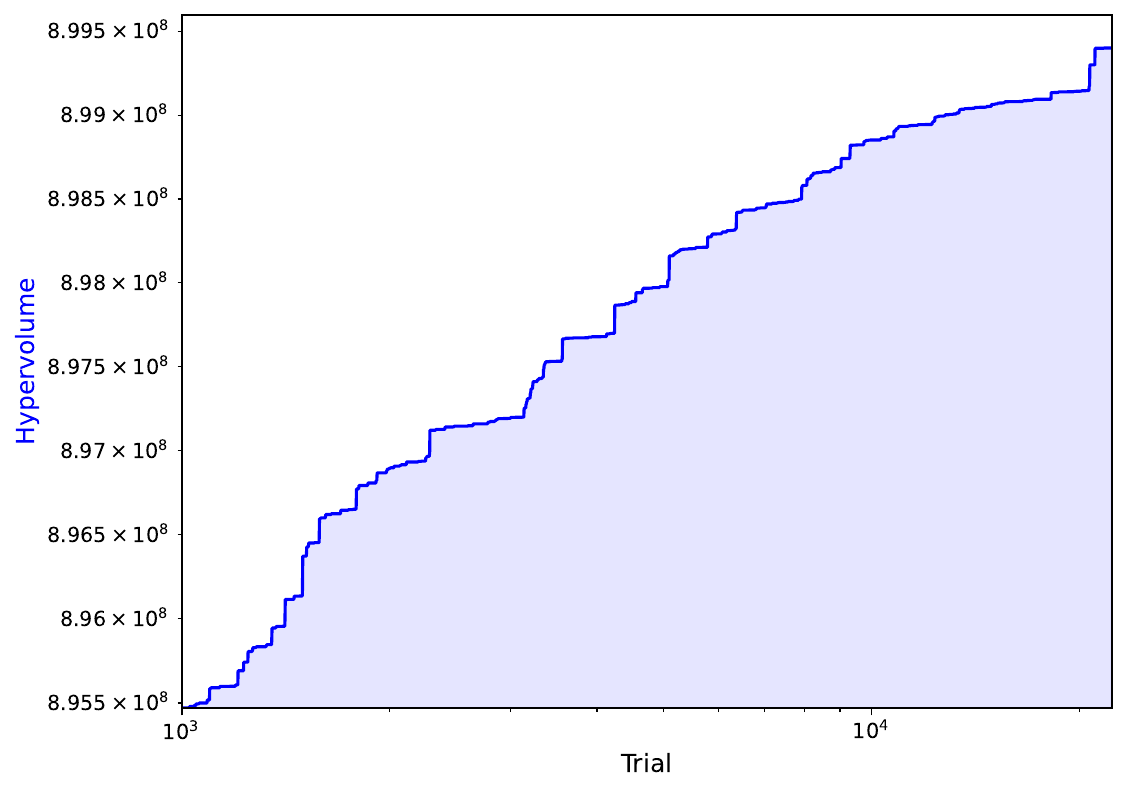}
    \caption{Hypervolume covered during optimization}
    \label{fig:hypervolume}
\end{figure}

\textbf{Hypervolume} is one of the key metrics used to evaluate the dynamics of
the Multi-Objective Bayesian Optimization (MOBO). The hypervolume measures the
extent of the objective space covered by the Pareto front, bounded by a
reference point, which, in this context, is the worst-case scenario for each
objective. A larger hypervolume indicates superior solution quality,
encompassing more objective space. This metric is also helpful as a convergence
test during the optimization process.The hypervolume increases and eventually
saturates in our experiments, as illustrated in the Figure
\ref{fig:hypervolume}. This saturation suggests that the optimization process
in our experiment has effectively converged to a stable set of optimal
solutions.

\begin{figure}[t]
    \includegraphics[width=\linewidth]{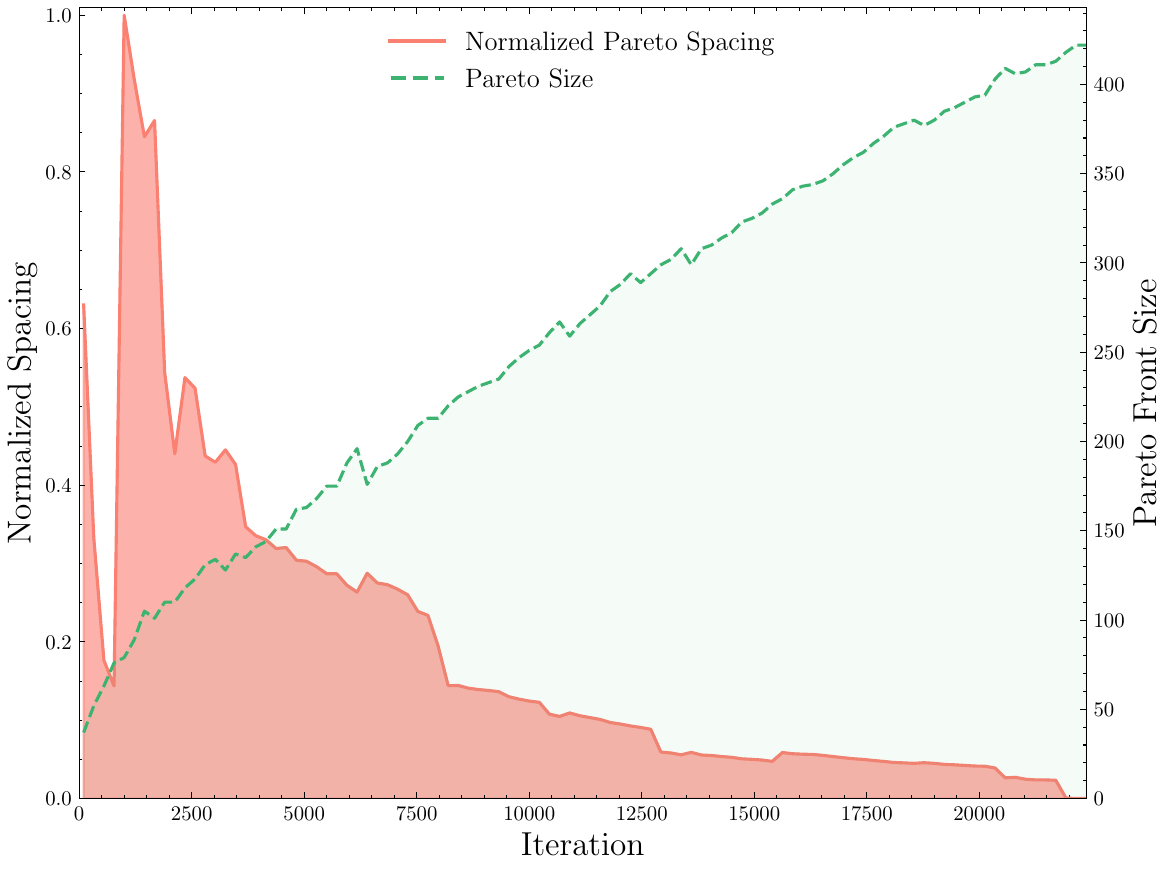}
    \caption{Spacing of Pareto points during optimization}
    \label{fig:pareto_spacing}
\end{figure}

\textbf{Pareto Diversity and Spacing:} Additionally, we use Pareto diversity
and spacing to characterize the effectiveness of our method. Pareto diversity
ensures that the optimization covers a broad range of trade-offs, enhancing
decision-making with diverse solutions. Meanwhile, spacing between these points
measures how uniformly solutions are distributed, indicating a balanced
solution space exploration. In Fig.~\ref{fig:pareto_spacing}, we observe the
evolution of these metrics in our experiment. The normalized Pareto spacing
decreases, showing an increase in uniformity among solutions. Meanwhile, the
Pareto size increases, indicating many optimal solutions. This shows that the
algorithm efficiently populates the objective space with well-distributed,
diverse solutions.

\subsection{Optimal Designs}

\begin{figure}[t]
    \includegraphics[width=\linewidth]{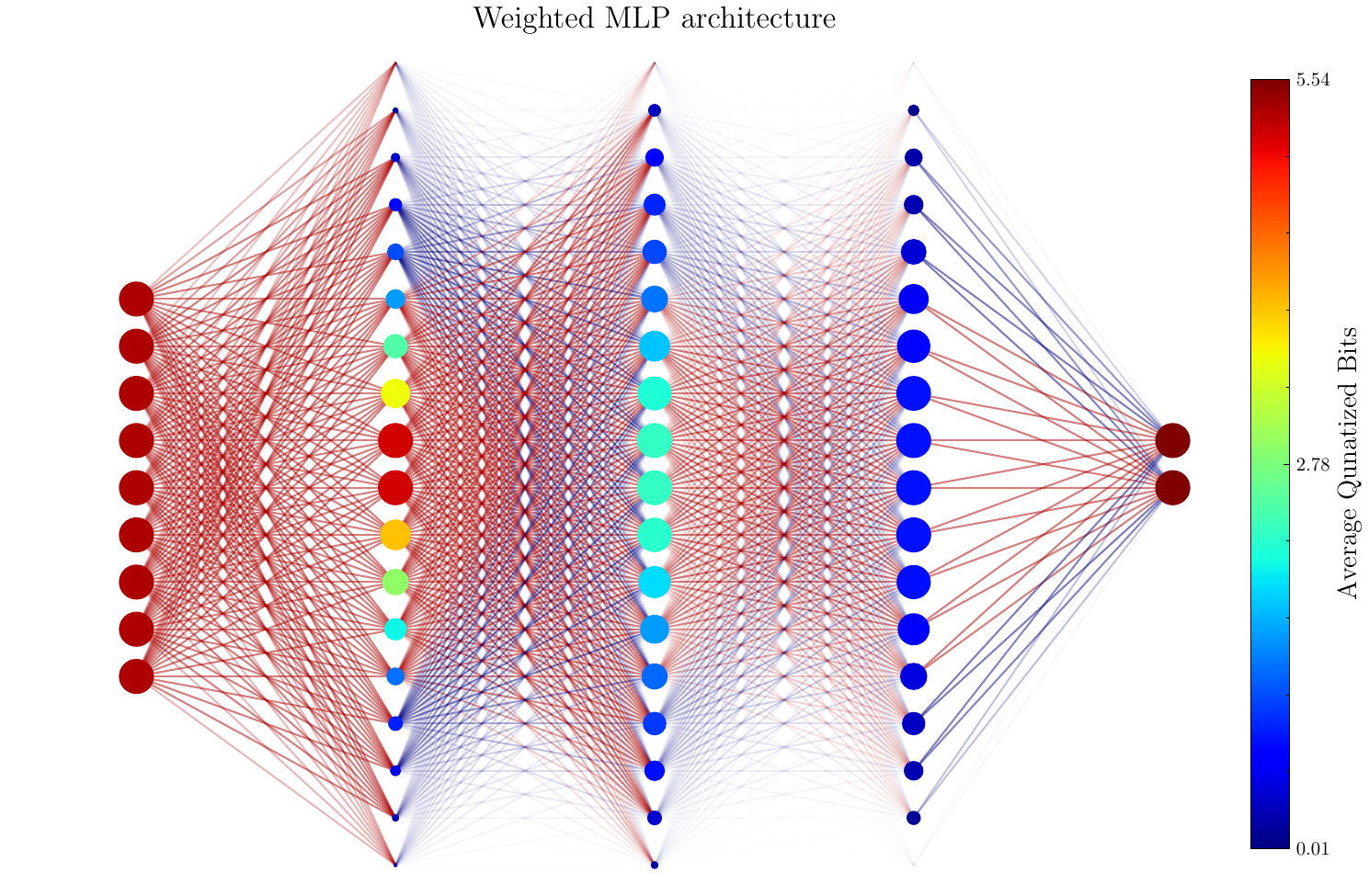}
    \caption{Average of quantization values in all MLP architectures at the
        Pareto front. Zero values used in points when a perception is absent in
        architecture.}
    \label{fig:optimal_mlp}
\end{figure}

The visual representation in Fig.~\ref{fig:optimal_mlp} represents an average
Multi-Layer Perceptron (MLP) architecture derived from a set of Pareto optimal
solutions from optimization in a co-design space with over two billion possible
configurations. This MLP model embodies the collective characteristics from
various configurations, differing depth, width, and quantization levels for
each layer. The size and color of the perceptron represent the average
quantization done in that perceptron in all optimal solutions with nodes that
are absent in a configuration represented with zero. The input-output
quantization is represented in the final layer of the MLP.
Fig.~\ref{fig:optimal_mlp} hints that the most detrimental effect of co-design
is lower quantization in input and output data.

\begin{figure}[t]
    \centering
    \includegraphics[width=0.8\linewidth]{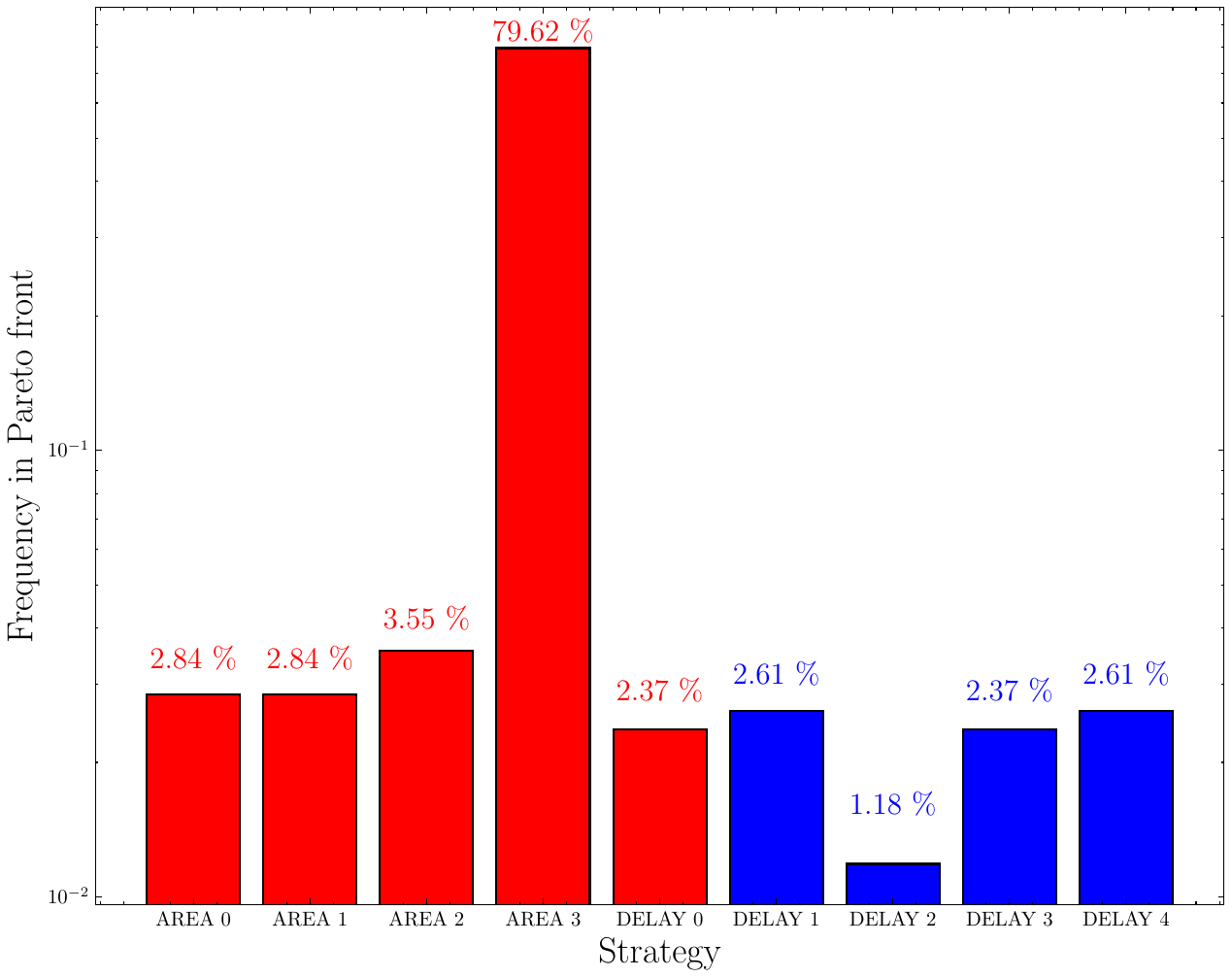}
    \caption{Distribution of the optimal synthesis strategy at the Pareto front.}
    \label{fig:optimal_strategies}
\end{figure}

Fig.~\ref{fig:optimal_strategies} presents the distribution of various
synthesis strategies at the Pareto front, effectively illustrating the
predominance of specific strategies over others in achieving optimal design
configurations.

Notably, the `AREA 3' strategy dominates the distribution, representing a
staggering 79.62\% of the choices at the Pareto front, indicating its
effectiveness in optimizing the design parameters within the constraints
provided. This could simply be because the optimization objective includes both
area and power, and AREA 3 is the most aggressive optimization for the area
and, hence, for power, too. A new simulation run with weighted values for area
and power can be used to confirm this hypothesis.

The other area strategies, each account for a relatively minor portion of the
Pareto front, all below 4\%, suggesting that while these strategies contribute
to the Pareto front, they are significantly less effective compared to "AREA
3". The significantly small role of DELAY 2 remains to be understood.

\subsection{Comparison with Theory} \label{sec:theory_vs_practice}

\begin{figure}
    \begin{subfigure}{\linewidth}
        \includegraphics[width=\linewidth]{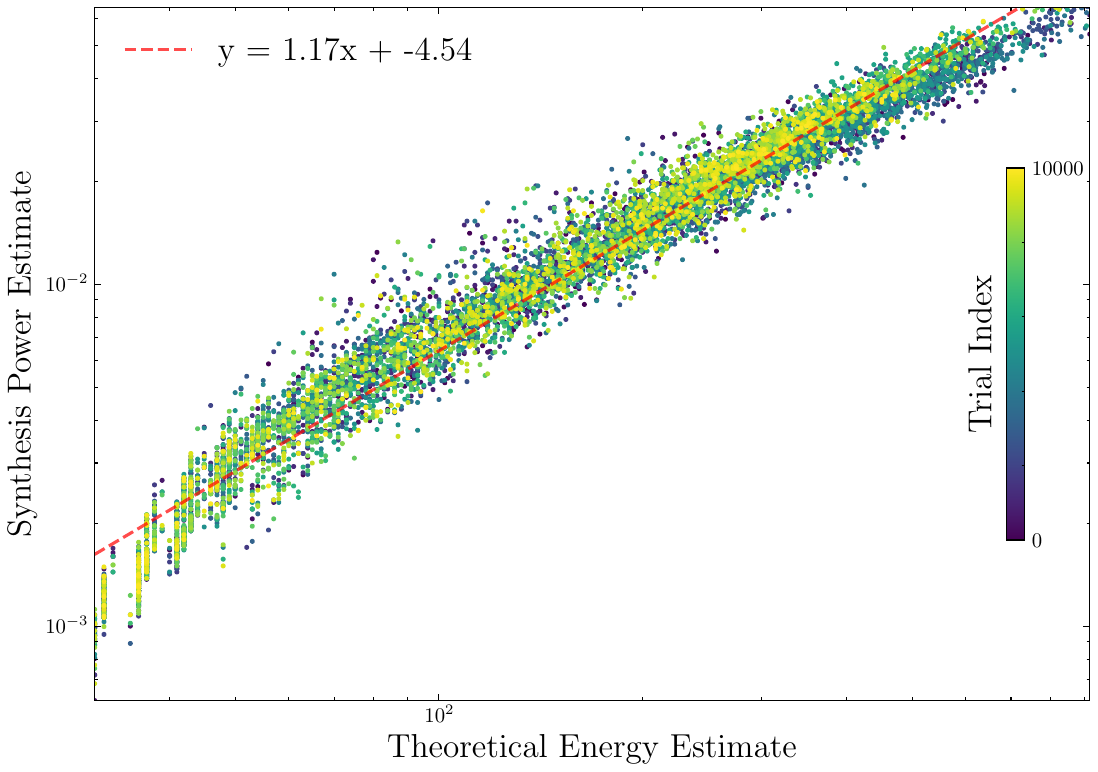}
        \caption{Correlation between theoretical estimate of energy consumption~\cite{horowitz20141} and the power estimated from synthesis report.}
        \label{fig:theory_correlation}
    \end{subfigure}
    \begin{subfigure}{\linewidth}
        \includegraphics[width=\linewidth]{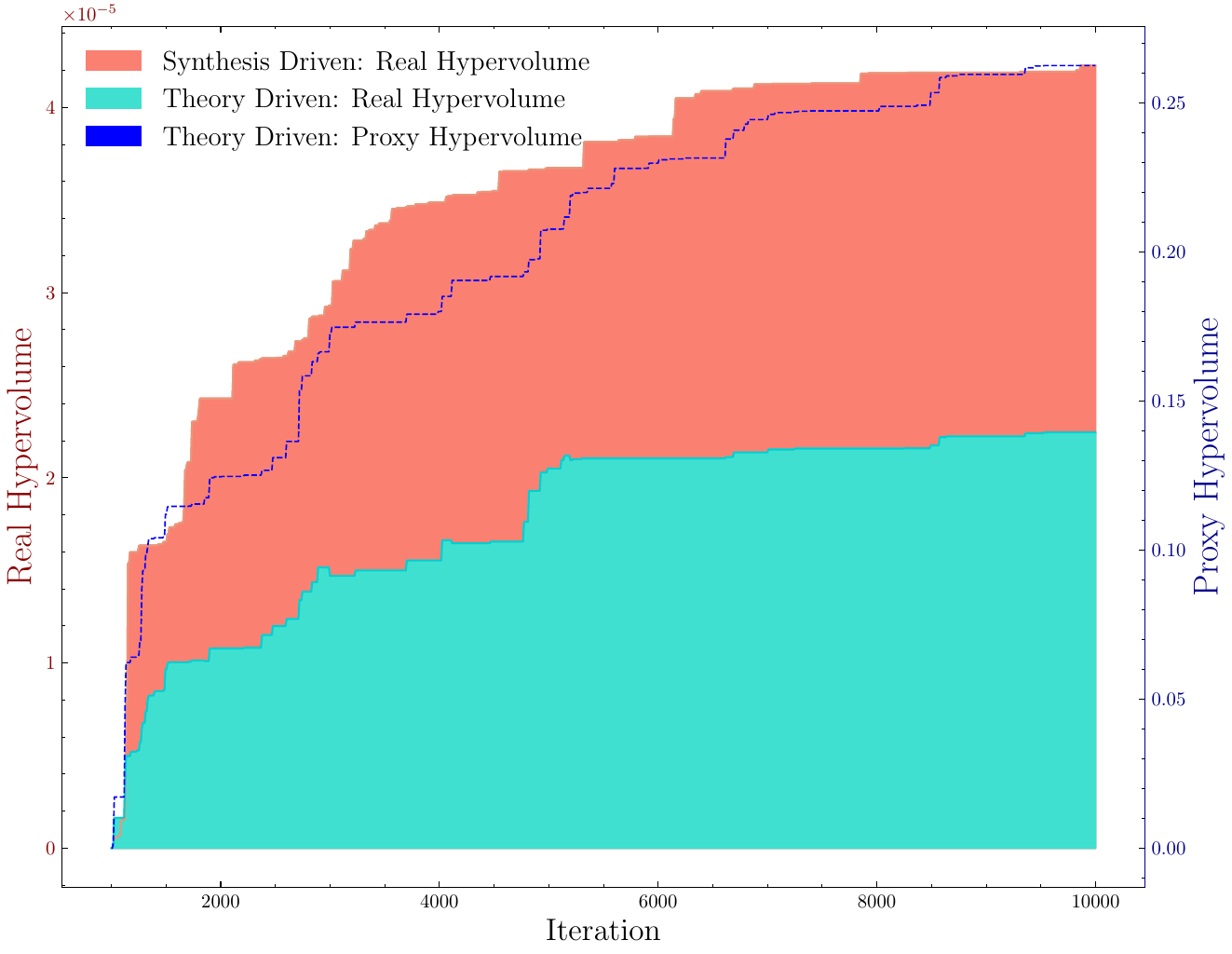}
        \caption{Evolution of hypervolume for a) power-val loss optimization using OpenLANE b) (dashed blue line) loss energy optimization using theory (Qkeras), referred in the figure as ``proxy hypervolume''. c) (greenish) The actual hypervolume being covered when optimization is driven by theoretical energy-estimate. }
        \label{fig:theory_hypervolume}
    \end{subfigure}
    \caption{Comparison of theoretical and circuit-derived energy estimates.}
\end{figure}

In this section, we do another experiment to quantify the distinction in our
approach, where we used a power estimate from ASIC synthesis instead of the
theoretical estimate. Fig.~\ref{fig:theory_correlation} shows that the
theoretical estimate of the energy shows a strong correlation with the power
estimated from the ASIC synthesis. Still, the uncertainty of its accuracy can
affect the search significantly. To quantify the effect of unquantified
uncertainty in the theoretical estimate, we ran another round of two-objective
optimization:a a) Area-Energy optimization guided by a theoretical estimate of
energy given by QKeras, and b) Power-energy optimization guided by power
estimated from logic synthesis.

To compare the effectiveness of the two approaches,
Fig.~\ref{fig:theory_hypervolume} shows the hypervolume evolution during the
two optimization runs, with a red-dashed line showing the proxy-hypervolume
used in theory-guided optimization. While the red curve shows an increase and
saturation like the synthesis-guided hypervolume, the plotting of `real
hypervolume' computed using power estimate from synthesis clearly shows that
theory-driven optimization saturates much earlier with very little Pareto
diversity. This empirical observation is the first insight into the negative
impact of theory-driven optimization for realistic ASIC design solutions, with
more in our future works.

\subsection{In-Pixel Implementable Neural Network}

We applied constraints on the trials to identify those suitable for in-pixel
implementation, as shown in Table~\ref{tab:parameters}. We chose the area
constraint of 250 x 250 $\mu$m$^2$ based on the 130nm process design kit (PDK)
used. The acceptable power density limit of 5 W/cm$^2$ was set to ensure
feasible on-chip power consumption. The delay constraint of 20 ps was derived
from the clock frequency used in our design. The validation Mean Squared Error
(MSE) threshold of 0.044837 corresponds to the MSE of a simple baseline model
that predicts the mean of the training data. These constraints collectively
define a challenging yet realistic design space for in-pixel neural network
implementation, pushing our co-design approach to find optimal solutions that
balance performance with hardware limitations.

\begin{table}[h]
    \centering
    \begin{tabular}{|l|l|}
        \hline
        \textbf{Parameter} & \textbf{Maximum Constraint} \\
        \hline
        Area               & $\leq$ 250 x 250 $\mu m^2$  \\
        \hline
        Power Density      & $\leq$ 5 W/cm$^2$           \\
        \hline
        Delay              & $\leq$ 20 ps                \\
        \hline
        Val. MSE           & $\leq$ 0.044837             \\
        \hline
    \end{tabular}
    \caption{Maximum design constraints for the in-pixel implementable Neural Network.}
    \label{tab:parameters}
\end{table}

\begin{figure}[t]
    \centering
    \includegraphics[width=\linewidth]{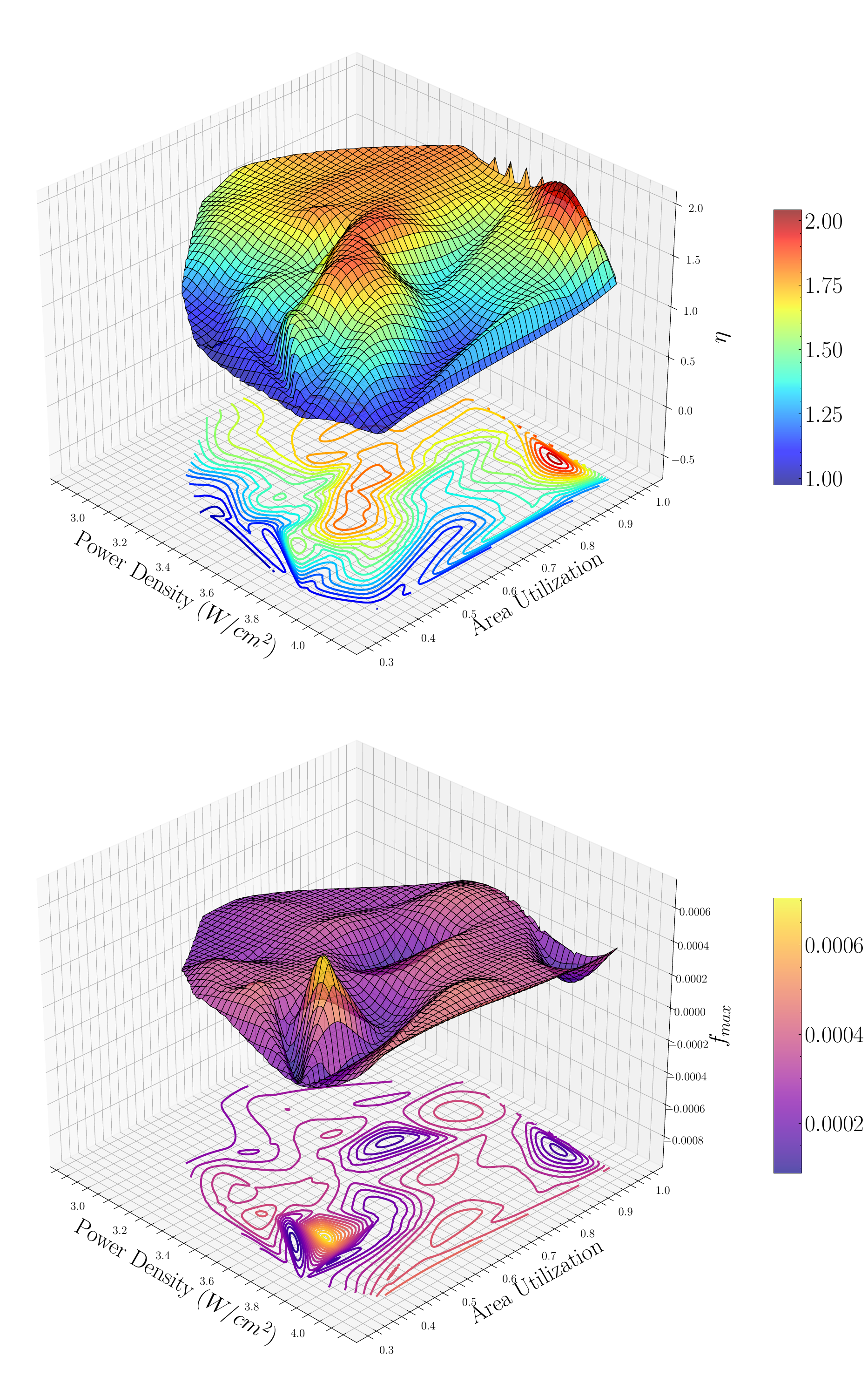}
    \caption{Optimized objectives at the Pareto front for the in-pixel implementable Neural Network.}
    \label{fig:in_pixel_pareto}
\end{figure}

Applying the constraints to the trials reveals a Pareto front with 54 in-pixel
implementable, equally good choices. Figure \ref{fig:in_pixel_pareto} plots the
optimization objectives in the constrained Pareto surface. The x-axis
represents the area utilization, which is the fraction of area occupied
compared to the constraint set. The y-axis shows the power density in W/cm².
The z-axis and color scale represent $\eta$, a performance metric of the model
defined as the ratio of the baseline MSE to the model's MSE (MSE of the
baseline / MSE of the model). The top plot illustrates the relationship between
area utilization, power density, and $\eta$. The surface shows a complex
landscape with multiple peaks and valleys, indicating the trade-offs between
these parameters. Generally, we can observe that higher $\eta$ values (better
performance) tend to correspond with higher power density and area utilization,
as expected for more complex models. The bottom plot focuses on
$f_{max}=\frac{1}{\text{Delay}}$, which represents the maximum achievable
frequency for each design point. This plot reveals a different pattern of
trade-offs, with peaks in $f_{max}$ occurring at various combinations of area
utilization and power density.

\begin{figure}[t]
    \centering
    \includegraphics[width=\linewidth]{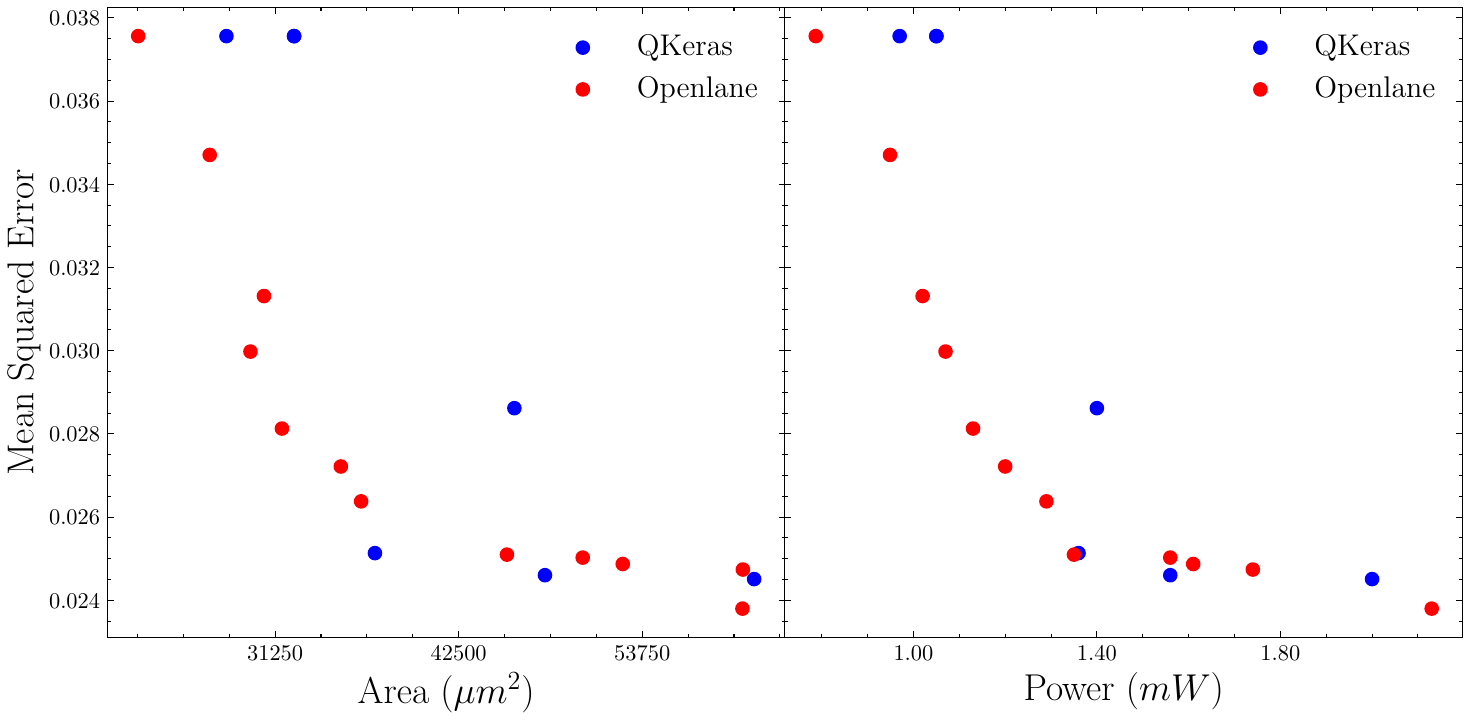}
    \caption{Comparison of the Pareto front between theory-guided (QKeras) vs
        simulation-guided (OpenLANE) after setting the in-pixel constraints.}
    \label{fig:theory_vs_practice_pareto}
\end{figure}

\begin{figure}[t]
    \centering
    \includegraphics[width=\linewidth]{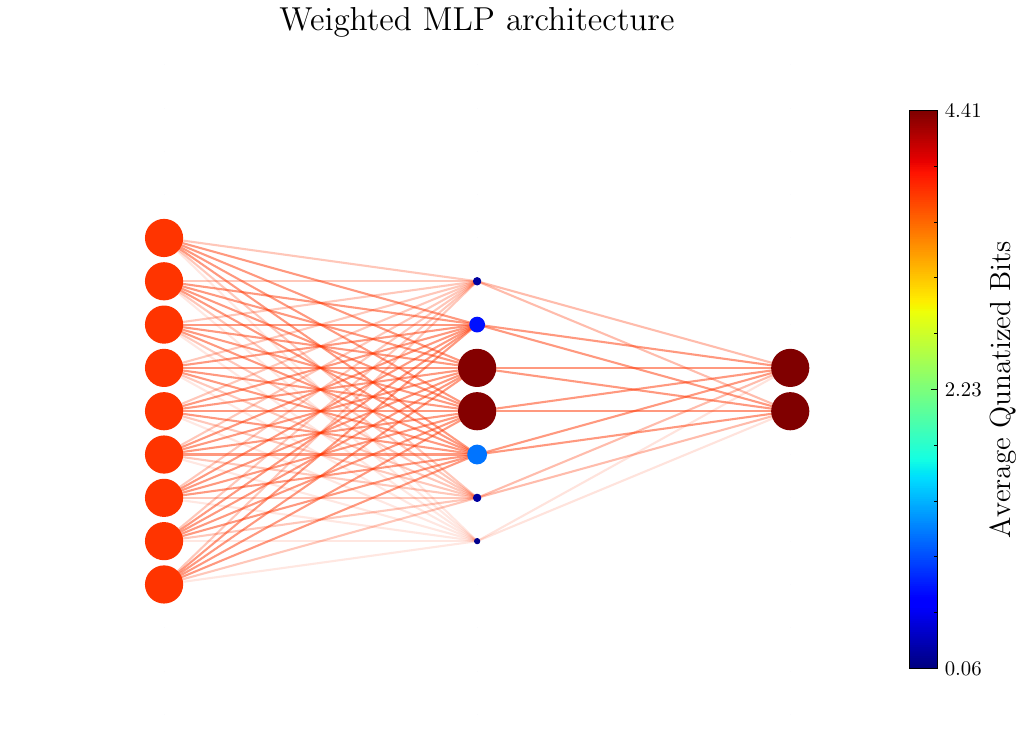}
    \caption{Average of quantization values in all MLP architectures at the
        Pareto front of in-pixel implementable NN. All models satisfying the
        constraints are single depth.}
    \label{fig:in_pixel_mlp}
\end{figure}

To further evaluate the effectiveness of using theoretical estimates for
guidance versus our approach using ASIC synthesis, we identified the
constrained Pareto fronts in the two-objective (performance and power)
optimization done in \ref{sec:theory_vs_practice}. The figure
\ref{fig:in_pixel_pareto} clearly demonstrates that utilizing synthesis-guided
metrics in the optimization process revealed a constrained Pareto front with a
greater number of in-pixel implementable design choices, each exhibiting better
area and power efficiency compared to those obtained using theoretical
estimate.

\section{Conclusion and Future Work}

In this paper, we presented a holistic approach to co-designing neural networks
and ASICs for enabling in-pixel intelligence in radiation detectors. Our key
contributions include an automated pipeline integrating neural network design
and ASIC synthesis within a multi-objective Bayesian optimization framework,
use of realistic ASIC synthesis metrics rather than theoretical estimates to
guide the optimization process for practical solutions, and exploration of a
specific large co-design space including neural architecture, quantization, and
ASIC synthesis strategies. We demonstrated Pareto-optimal designs meeting
strict area, power, and latency constraints for in-pixel implementation.

Our results show the benefits of using circuit-level metrics from ASIC
synthesis compared to theoretical estimates, which can lead to less ideal
solutions even in cases when such approximations are available. This approach
provides more accurate guidance towards implementable designs.

There are several promising directions for future work. Expanding the co-design
space to include additional neural architectures beyond MLPs could yield more
powerful in-pixel processing capabilities. Incorporating more detailed
circuit-level optimization choices into the co-design process may uncover
additional efficiency gains. Exploring more emerging technologies could push
the boundaries of what's possible with in-pixel processing. Finally, extending
this approach to other scientific applications requiring extreme edge AI could
have broad impact across various fields.

\section{Acknowledgement}

This manuscript has been authored by Brookhaven Science Associates, LLC under
Contract No. DE-SC0012704 with the U.S. Department of Energy. The United States
Government retains and the publisher, by accepting the article for publication,
acknowledges that the United States Government retains a non-exclusive,
paid-up, irrevocable, world-wide license to publish or reproduce the published
form of this manuscript, or allow others to do so, for United States Government
purposes. This work was supported by the Laboratory Directed Research and
Development (LDRD) Program of Brookhaven National Laboratory under U.S.
Department of Energy Contract No. DE-SC0012704. The authors would also like to
thank Dr. Sandeep Miryala and Dr. Sanket Jantre for numerous helpful
discussions.

\bibliographystyle{acm}
\bibliography{references}

\clearpage

\end{document}